\documentclass[11pt,a4paper]{article}
\usepackage[hyperref]{emnlp2020}
\usepackage{times}
\usepackage{latexsym}
\usepackage{soul}
\usepackage{url}

\usepackage{graphicx}
\usepackage{amsmath}
\usepackage{amssymb}
\usepackage{amsthm}
\usepackage{booktabs}
\usepackage{algorithm}
\usepackage{algorithmic}
\usepackage{boldline}
\usepackage{tikz}
\usetikzlibrary{shapes,arrows}
\usepackage{caption}
\usepackage{subcaption}
\usepackage{tikz-dependency}
\usepackage{wrapfig}
\usepackage{pgfplots}
\usepackage{filecontents}
\usepackage[T1]{fontenc}
\usepackage{subfiles}
\usepackage{readarray}
\usepackage{xcolor}
\usepackage{import}
\usepackage{multirow}
\usepackage{hhline}
\usepackage{mathtools}

\usepackage{microtype}

\aclfinalcopy % Uncomment this line for the final submission
\def\aclpaperid{1615} %  Enter the acl Paper ID here

\newcommand{\argmax}{\operatornamewithlimits{argmax}}

\newcommand{\xvec}{\mathbf{x}}
\newcommand{\yvec}{\mathbf{y}}

\newcommand{\vvec}{\mathbf{v}}
\newcommand{\Wvec}{\mathbf{W}}
\newcommand{\Uvec}{\mathbf{U}}

\newcommand{\rvec}{\mathbf{r}}

\newcommand{\mcL}{\mathcal{L}}

\title{AIN: Fast and Accurate Sequence Labeling \\with Approximate Inference Network}

\author{\parbox{\linewidth}{\centering Xinyu Wang$^{\diamond\ddagger}$, Yong Jiang$^{\dagger}$\textsuperscript{$\ast$}, Nguyen Bach$^{\dagger}$, \\ Tao Wang$^{\dagger}$, Zhongqiang Huang$^{\dagger}$, Fei Huang$^{\dagger}$,  Kewei Tu$^{\diamond}$\thanks{\hspace{1mm} Yong Jiang and Kewei Tu are the corresponding authors. $^{\ddagger}$: This work was conducted when Xinyu Wang was interning at Alibaba DAMO Academy. $^{\diamond}$: {\tt \{wangxy1, tukw\}@shanghaitech.edu.cn}, $^{\dagger}$: {\tt \{yongjiang.jy, nguyen.bach, leeo.wangt, z.huang, f.huang\}@alibaba-inc.com}}} \\
 $^\diamond$School of Information Science and Technology, ShanghaiTech University \\
 $^{\diamond}$Shanghai Engineering Research Center of Intelligent Vision and Imaging \\
 $^{\diamond}$University of Chinese Academy of Sciences \\
 $^\dagger$DAMO Academy, Alibaba Group \\
 }

\date{}

\begin{document}
\maketitle

\begin{abstract}
%with pretrained word embeddings and contextual feature extractors such as RNN or CNN 
The linear-chain Conditional Random Field (CRF) model is one of the most widely-used neural sequence labeling approaches. Exact probabilistic inference algorithms such as the forward-backward and Viterbi algorithms are typically applied in training and prediction stages of the CRF model. However, these algorithms require sequential computation that makes parallelization impossible. In this paper, we propose to employ a parallelizable approximate variational inference algorithm for the CRF model. Based on this algorithm, we design an approximate inference network that can be connected with the encoder of the neural CRF model to form an end-to-end network, which is amenable to parallelization for faster training and prediction. The empirical results show that our proposed approaches achieve a 12.7-fold improvement in decoding speed with long sentences and a competitive accuracy compared with the traditional CRF approach.

\end{abstract}

\section{Introduction}
Sequence labeling assigns each token with a label in a sequence. Tasks such as Named Entity Recognition (NER) \cite{Sundheim1995NamedET}, Part-Of-Speech (POS) tagging \cite{derose-1988-grammatical} and chunking \cite{tjong-kim-sang-buchholz-2000-introduction} can all be formulated as sequence labeling tasks. BiLSTM-CRF \cite{huang2015bidirectional,lample-etal-2016-neural,ma-hovy-2016-end} is one of the most successful neural sequence labeling architectures. It feeds pretrained (contextual) word representations into a single layer bi-directional LSTM (BiLSTM) encoder to extract contextual features and then feeds these features into a CRF \cite{10.5555/645530.655813} decoder layer to produce final predictions. The CRF layer is a linear-chain structure that models the relation between neighboring labels. In the traditional CRF approach, exact probabilistic inference algorithms such as the forward-backward and Viterbi algorithms are applied for training and prediction respectively. 
%The Viterbi algorithm is applied to exactly find the best label sequence in inference and the forward-backward algorithm is applied to compute posterior marginal distributions exactly for each position in training. 
In many sequence labeling tasks, the CRF layer leads to better results than the simpler method of predicting each label independently.

In practice, we sometimes require very fast sequence labelers for training (e.g., on huge datasets like WikiAnn \cite{pan-etal-2017-cross}) and prediction (e.g. for low latency online serving). The BiLSTM encoder and the CRF layer both contain sequential computation and require $O(n)$ time over $n$ input words even when parallelized on GPU. A common practice to improve the speed of the encoder is to replace the BiLSTM with a CNN structure \cite{collobert2011natural,strubell-etal-2017-fast}, distill larger encoders into smaller ones \cite{tsai-etal-2019-small,mukherjee2020tinymbert} or in other settings \cite{tu2018learning,yang-etal-2018-design,tu-gimpel-2019-benchmarking,cui-zhang-2019-hierarchically}. The CRF layer, however, is more difficult to replace because of its superior accuracy compared with faster alternatives in many tasks. %More recently, \citet{cui-zhang-2019-hierarchically} proposed BiLSTM-LAN to replace the CRF layer, which has a lower time complexity, but the network introduces 3 additional LSTM layers that require sequential computations. The CRF layer is still necessary for better accuracy in many tasks, which limits the speed.
% \cite{zheng2015conditional} showed such an algorithm can be unfolded as an RNN on grid-structure, we expand the work on the sequence structure and unfold the MFVI algorithm as an RNN as will

In order to achieve sublinear time complexity on the CRF layer, we must parallelize the CRF prediction over the tokens. In this paper, we apply Mean-Field Variational Inference (MFVI) to approximately decode the linear-chain CRF. MFVI iteratively passes messages among neighboring labels to update their distributions locally. Unlike the exact probabilistic inference algorithms, MFVI can be parallelized over different positions in the sequence, achieving time complexity that is constant in $n$ with full parallelization.
%Similar to \citet{zheng2015conditional}, we show that such an algorithm can be unfolded as an RNN, 
Previous work \cite{zheng2015conditional} showed that such an algorithm can be unfolded as an RNN for grid CRF structure. We expand on the work for the linear-chain CRF structure and unfold the algorithm as an RNN
which can be connected with the encoder to form an end-to-end neural network that is amenable to parallelization for both training and prediction. We call the unfolded RNN an approximate inference network (AIN). In addition to linear-chain CRFs, we also apply AIN to factorized second-order CRF models, which consider relations between more neighboring labels. Our empirical results show that AIN significantly improves the speed and achieves competitive accuracy against the traditional CRF approach on 4 tasks with 15 datasets.

\section{Approaches}
Given an input sequence with $n$ tokens $\xvec=[x_1,x_2,\dots,x_n]$ and a corresponding label sequence $\yvec=[y_1,y_2,\dots,y_n]$ with a label set of size $L$, the conditional probability of $\yvec$ given $\xvec$ specified by a CRF with position-wise factorization is:
\begin{align}
% \psi(\xvec,\yvec)&=\sum\limits_{i=1}^{n}\psi(\xvec, \yvec, i) \nonumber\\
P(\yvec|\xvec) &= \frac{ \exp\{\sum\limits_{i=1}^{n}\psi(\xvec, \yvec, i)\}}{\sum\limits_{\yvec^{\prime} \in \mathcal{Y}(\xvec)} \exp\{\sum\limits_{i=1}^{n}\psi(\xvec, \yvec^{\prime}, i))\}}\nonumber%\label{eq:sentprob}
\end{align}
where $\mathcal{Y}(\mathbf{x})$ is the set of all possible label sequences for $\xvec$ and $\psi(\xvec, \yvec, i)$ is a potential function.

% \begin{figure}[t]
% \centering
% \includegraphics[width=0.49\textwidth]{model_struct.pdf}
% \caption{Illustration of our approximate inference network.}
% \label{fig:model_structure}
% \end{figure}

\begin{figure}[t]
\centering
\includegraphics[width=0.47\textwidth]{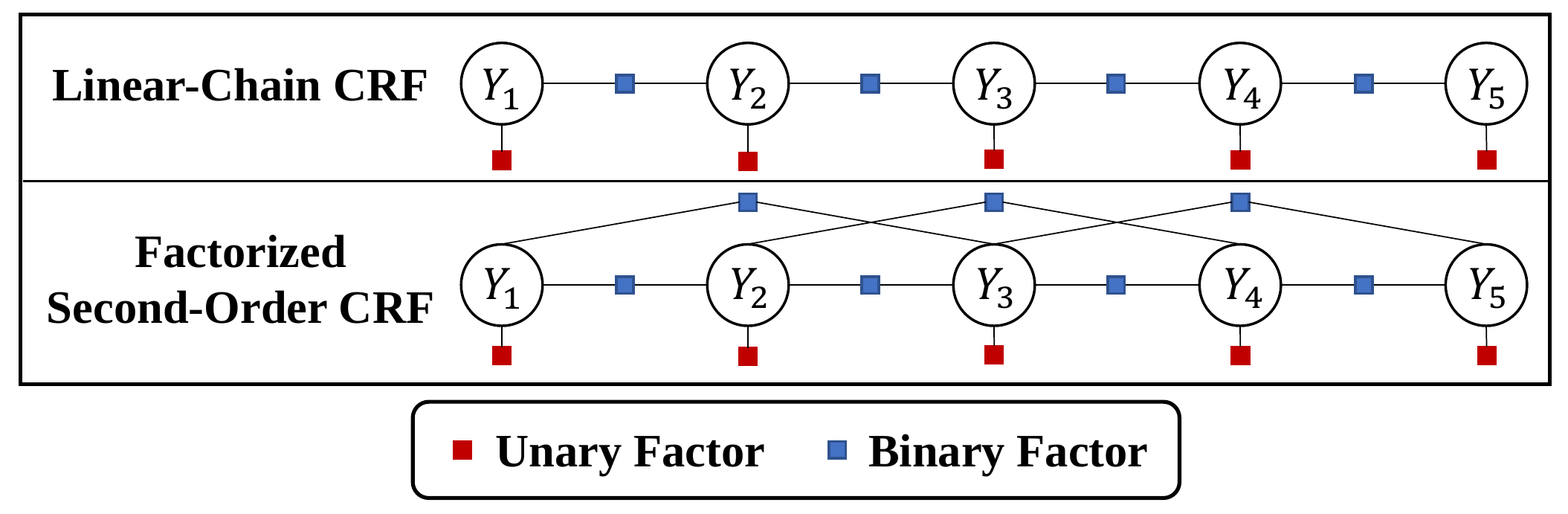}
\caption{Factor graphs of different CRFs. $Y_i$ is the random variable representing the $i$-th label.}
\label{fig:crf_structure}
\end{figure}

In the simplest case, the potential function is just a softmax function that outputs the distribution of each label independently. We call it the MaxEnt approach. In a typical linear-chain CRF, the potential function is decomposed into a unary potential $\psi_u$ and a binary potential $\psi_b$ (called the emission and transition functions respectively):
\begin{align}
&\psi(\xvec, \yvec, i) = \psi_u(\xvec, y_i) + \psi_b(y_{i-1}, y_i) \label{eq:crf_potential} \\
&\psi_u(\xvec, y_i) = \rvec_i\Wvec \vvec_{y_i} \nonumber\\
&\psi_b(y_{i-1}, y_i) = \Uvec_{y_{i-1},y_i} \label{eq:binary}
\end{align}
where $\rvec_i$ is the contextual feature of $x_i$ output from the CNN or BiLSTM encoder with dimension $d$, $\vvec_{y_i}$ is a one-hot vector for label $y_i$, $\Wvec$ is a $d \times L$ matrix and $\Uvec$ is an $L \times L$ matrix containing the transition scores between two labels. The factor graph of a linear-chain CRF is shown at the top of Figure \ref{fig:crf_structure}.

The exact probabilistic inference algorithms (Viterbi and forward-backward)
%The Viterbi and forward-backward algorithms 
for the CRF layer are significantly slower than the MaxEnt approach. They take $O(nL^2)$ and $O(n\log L)$ time on CPU and GPU\footnote{We assume that the number of threads is enough for full parallelization on GPU and the parallel reduction (e.g., sum and max) for a $L$ elements vector takes $O(\log L)$ time \cite{harris2007optimizing}.} respectively, while the decoder in MaxEnt takes $O(nL)$ and $O(\log L)$.

\subsection{AIN on Linear-Chain CRF}
\label{sec:first-order}
In order to speed up the training and prediction time of the CRF layer, we propose the approximate inference network (AIN), which is a neural network derived from MFVI for approximate decoding in linear-chain CRF.

MFVI approximates the distribution $P(\yvec|\xvec)$ with a factorized distribution $Q(\yvec|\xvec)=\prod\limits_{i=1}^{n} Q_i(y_i|\xvec)$ and update it iteratively to minimize the KL divergence $KL(Q||P)$. The update formula of $Q_i(y_i|\xvec)$ at iteration $m$ is:
% \begin{align}
% Q_i^{m}(y_i|\xvec) \propto& \exp\{\psi_u(\xvec, y_i) \nonumber\\
% &+ \sum\limits_{j\in \mcN(i)}\sum\limits_{y_j\in Y_j}Q^{m-1}_j(y_j)\psi_b(y_i, y_j)\} \label{eq:mf_1st}
% \end{align}
\begin{align*}
s(i,j,k){:=}\sum\limits_{\mathclap{y_{i}=1}}^L Q^{k{-}1}_{i}(y_{i}|\xvec)\psi_b(y&_{\min\{i,j\}}, y_{\max\{i,j\}})\\
Q_i^{m}(y_i|\xvec) {\propto} \exp\{\psi_u(\xvec, y_i)+&s(i{-}1,i,m)\nonumber\\
{+}s&(i{+}1,i,m)\} \nonumber
\end{align*}
where $s(i,j,k)$ represents the message from node $i$ to node $j$ at time step $k$. $Q_i^{0}(y_i|\xvec)$ is set by normalizing the unary potential $\psi_u(\xvec, y_i)$. Upon convergence, the label sequence with the highest approximate probability $Q(\yvec|\xvec)$ can be found by optimizing $Q_i (y_i|\xvec)$ at each position $i$:
\begin{align}
\hat{y_i} &= \argmax \limits_{y_i \in \{1,\dots,L\}} Q_i(y_i|\xvec) \nonumber
\end{align}

Similar to \citet{zheng2015conditional}, we unfold the MFVI algorithm as a recurrent neural network that is parameterized by the linear-chain CRF potential functions. We fix the number of iterations to $M$ and call the resulting network AIN. AIN can be connected with the encoding network that computes the potential functions and together they form an end-to-end neural network. 

Note that, different from previous work \cite{NIPS2011_4296,zheng2015conditional,baque2016principled,7913730,wang-etal-2019-second} using the MFVI algorithm for solving intractable problems of densely connected probabilistic models to get better accuracy, we propose to employ the MFVI algorithm to accelerate tractable inference of sequence-structured probabilistic models. As far as we know, this is the first attempt of using approximate inference on tractable models for speedup with GPU parallelization.

\begin{figure}[t]
\centering
\includegraphics[width=0.4\textwidth]{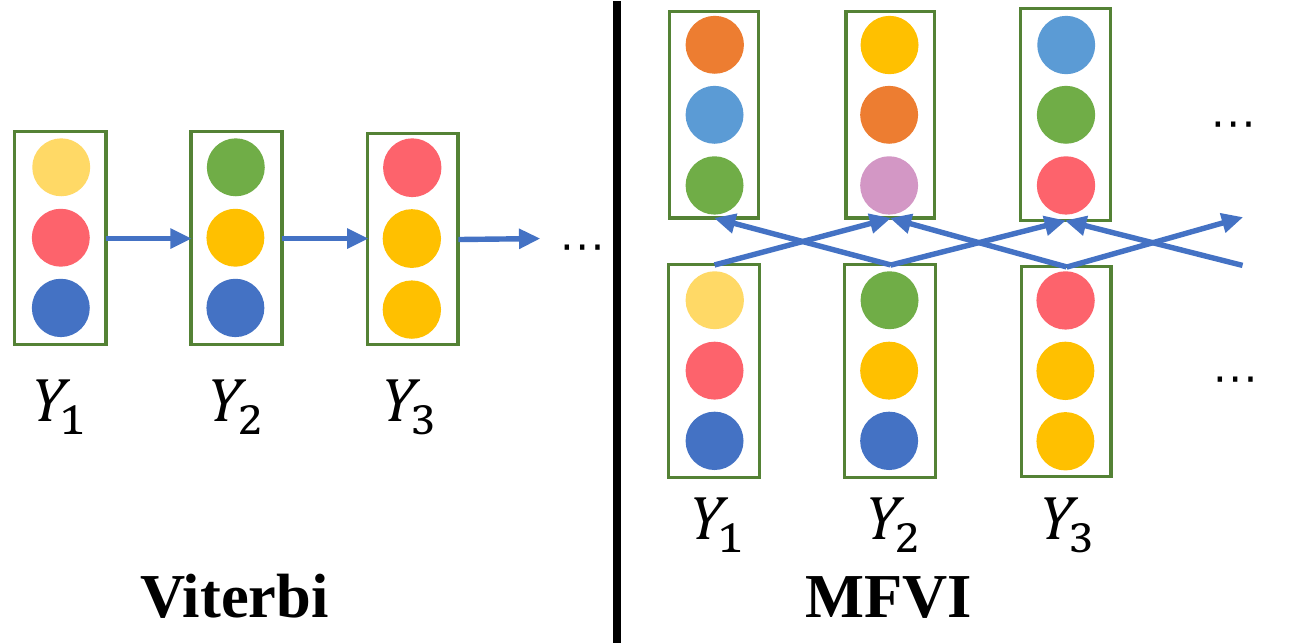}
\caption{Illustration of the computation graphs for the Viterbi decoding and one iteration of our MFVI inference on the CRF model. $Y_i$ is the random variable representing the $i$-th label with three possible values. The illustrated vectors represent Viterbi scores and $Q_i$ distributions respectively.}
\label{fig:inf_compare}
\end{figure}

The time complexity of each iteration of the MFVI algorithm is $O(nL^2)$, which is on par with the time complexity of the exact probabilistic inference algorithms. However, in each iteration, the update of each distribution $Q_i(y_i|\xvec)$ depends only on its two neighboring distributions from the previous iteration, so each iteration can be parallelized over positions. A comparison between the Viterbi algorithm and the MFVI algorithm is shown in Figure \ref{fig:inf_compare}. The time complexity of our AIN decoder with full GPU parallelization is $O(M \log L)$, while the time complexity of the exact probabilistic inference algorithms with GPU parallelization is $O(n \log L)$. We set the value of $M$ to $3$s , which is much smaller than the typical value of sequence length $n$.

\subsection{AIN on Factorized Second-Order CRF}
\label{sec:second-order}

We can extend AIN to the second-order CRF with a ternary potential function over every three consecutive labels. In the second-order CRF, the potential function in Eq. \ref{eq:crf_potential} becomes:
% \begin{align}
% \psi(\xvec, \yvec, i) &= \psi_u(\xvec, y_i) + \sum\limits_{j\in \mcT(i)}\sum\limits_{k\in \mcT(i)}\psi_t(y_i, y_j, y_k) \nonumber%\label{eq:mf_2nd_potential} 
% \end{align}
\begin{align}
\psi(\xvec, \yvec, i) &= \psi_u(\xvec, y_i) + \psi_t(y_{i-2}, y_{i-1}, y_{i}) \nonumber%\label{eq:mf_2nd_potential} 
\end{align}
However, the second-order CRF has space and time complexity that is cubic in $L$. Therefore, we factorize its ternary potential function and reduce its complexity to be quadratic in $L$:
\begin{align}
\psi_t(y_{i-2}, y_{i-1}, y_{i}) &=\psi_{b^{\prime}}(y_{i-2},y_i) + \psi_b(y_{i-1},y_i) \nonumber \\
\psi_{b^{\prime}}(y_{i-2},y_i) &= \Tilde{\Uvec}_{y_{i-2},y_i}
\nonumber
% \label{eq:mf_ternary} 
\end{align}
where the matrix $\Tilde{\Uvec}$ has the same shape as $\Uvec$ in Eq. \ref{eq:binary}. The factor graph of our factorized second-order CRF is shown at the bottom of Figure \ref{fig:crf_structure}. The update formula is similar to that of our first-order approach but with more neighbors:
\begin{align*}
&Q_i^{m}(y_i|\xvec) {\propto} \exp\{\psi_u(\xvec, y_i){+}s^{\prime}(i{-}2,i,m)\\
&{+}s(i{-}1,i,m){+}s(i{+}1,i,m){+}s^{\prime}(i{+}2,i,m)\} 
\end{align*}
where $s^{\prime}(i,j,k)$ has a similar definition as $s(i,j,k)$ by replacing $\psi_b$ with $\psi_{b^{\prime}}$. The time complexity of this approach is also $O(nL^2)$ for each iteration and $O(M \log L)$ with full GPU parallelization for $M$ iterations.
Following the first approach, we also unfold MFVI of this approach as an AIN.

\subsection{Learning}
Given a sequence $\xvec$ with corresponding gold labels $\yvec^* = \{y_1^*, \cdots, y_n^*\}$, the learning objective of our approaches is:
\begin{align*}
\mcL_{\text{NLL}} = - \sum\limits_{i=1}^{n} \log Q_{i}^{M}(y^*_{i}|\xvec)
\end{align*}
Since AINs are end-to-end neural networks, the objective function can be optimized by any gradient-based method in an end-to-end manner.
%--------------------------------------Tables------------------------------------

\begin{table*}[ht!]
\centering
\small
% \begin{tabular}{l||r|r|r|r|r|r|r|r||r|r|r|r||r}
\begin{tabular}{l||cccc|cccc||cc|cc}
\hlineB{4}
 & \multicolumn{8}{c||}{\bf \textsc{Word-Char-BiLSTM}} & \multicolumn{4}{c}{\bf \textsc{Word-CNN}}\\
\hhline{~||------------}
\hhline{~||------------}
 & \multicolumn{4}{c|}{Training}                 & \multicolumn{4}{c||}{Prediction} & \multicolumn{2}{c|}{Training}                 & \multicolumn{2}{c}{Prediction}\\
\hhline{-||------------}
\hhline{-||------------}
\# Words & \multicolumn{2}{c}{32}                 & \multicolumn{2}{c|}{128}        & \multicolumn{2}{c}{32} & \multicolumn{2}{c||}{128} & \multicolumn{1}{c}{32} & \multicolumn{1}{c|}{128} & \multicolumn{1}{c}{32} & \multicolumn{1}{c}{128}\\
 %\multicolumn{2}{c|}{32 words} & \multicolumn{2}{c||}{128 words} &\\
\hhline{-||------------}
\hhline{-||------------}
 & All & Dec. & All & Dec. & All & Dec. & All & Dec. & All & All & All & All\\
\hline\hline
MaxEnt$^{\star}$ & 6.8$\times$ & -           & 13.1$\times$ & -            & 3.0$\times$ & -           & 5.9$\times$ & -            & 12.9$\times$ & 40.1$\times$ & 6.3$\times$ & 18.6$\times$\\
\hline
AIN-1O   & 4.3$\times$ & 7.7$\times$ & 10.2$\times$ & 31.4$\times$ & 1.7$\times$ & 2.4$\times$ & 4.4$\times$ & 12.7$\times$ & 5.6$\times$  & 21.5$\times$ & 2.4$\times$ & 6.8$\times$\\
% AIN-2   & 3.4$\times$ & 5.0$\times$ & 8.6$\times$  & 19.3$\times$ & 1.4$\times$ & 1.8$\times$ & 4.1$\times$ & 10.1$\times$  & 4.3$\times$  & 16.2$\times$ & 1.7$\times$ & 5.3$\times$  & 3.43 \\
AIN-F2O  & 3.5$\times$ & 5.3$\times$ & 8.7$\times$  & 20.1$\times$ & 1.5$\times$ & 1.9$\times$ & 4.1$\times$ & 10.6$\times$  & 4.4$\times$  & 16.7$\times$ & 1.8$\times$ & 5.5$\times$\\
\hlineB{4}
\end{tabular}
\caption{Relative speedup over the \textbf{CRF} model with 10,000 sentences of 32/128 words. \textbf{All} represents the speed of the full model. \textbf{Dec.} represents the speed of decoder. $\star$: For reference.}
\label{tab:speeds}
\end{table*}

\begin{table*}[ht!]
\setlength\tabcolsep{2.5pt}
\small
\centering
\begin{tabular}{l||cccc|c||cccc|c||cccc|c}
\hlineB{4}
% \midrule
% \hline
& \multicolumn{5}{c||}{\bf \textsc{Word-Char-BiLSTM}} & \multicolumn{5}{c||}{\bf \textsc{Word-CNN}}&\multicolumn{5}{c}{\bf \textsc{Word Only}}\\
\hhline{~||-----||-----||-----}
% \midrule 
  & NER & POS & Chunk & SF & Avg. & NER & POS & Chunk & SF & Avg. & NER & POS & Chunk & SF & Avg. \\
\hline
\hline
% \midrule 
MaxEnt$^{\star}$ & 83.74 & 94.84 & 92.58 & 95.47 & 91.65 & 75.19 & 94.00 & 87.05 & 91.07 & 86.83 & 52.27 & 90.53 & 78.17 & 62.93 & 70.98 \\
\hhline{-||----|-||----|-||----|-}
% \midrule 
CRF & 84.17 & 94.91 & \textbf{92.88} & 95.52 & 91.87 & \textbf{79.44} & 94.26 & \textbf{89.21} & 92.24 & \textbf{88.79} & \textbf{72.28} & 92.79 & \textbf{89.39} & 76.82 & 82.82 \\
AIN-1O & \textbf{84.22} & \textbf{94.97} & 92.87 & \textbf{95.59} & \textbf{91.91} & 78.47 & 94.29 & 88.86 & 92.18 & 88.45 & 70.23 & 92.84 & 88.69 & 88.76 & 85.13 \\
AIN-F2O & 84.11 & 94.91 & 92.85 & 95.58 & 91.86 & 78.71 & \textbf{94.32} & 88.75 & \textbf{92.26} & 88.51 & 71.16 & \textbf{93.03} & 88.80 & \textbf{88.86} & \textbf{85.46} \\
\hlineB{4}

\end{tabular}
\caption{Averaged F1 score and accuracy on four tasks. \textbf{SF} represents the slot filling task. $\star$: For reference.}
\label{tab:ner_pos}
\end{table*}

%--------------------------------------Tables------------------------------------

\section{Experiments}

% \subsection{Datasets}
% \paragraph{Named Entity Recognition (NER)} We use the corpora from the CoNLL 2002 and CoNLL 2003 shared tasks \cite{tjong-kim-sang-2002-introduction,tjong-kim-sang-de-meulder-2003-introduction}, which contain four languages in total. We use the standard training/development/test split for experiments.
% \paragraph{Chunking} The chunking datasets are also from the CoNLL 2003 shared task \cite{tjong-kim-sang-de-meulder-2003-introduction} that contains English and German datasets. We use the same standard split as in NER.
% \paragraph{POS Tagging} Universal Dependencies (UD) \cite{11234/1-2837} contains syntactically annotated corpora of over 70 languages. We use universal POS tag annotations with 8 languages for experiments. 
% \paragraph{Slot Filling} Slot filling is a task that interprets user commands by extracting relevant slots, which can be formulated as a sequence labeling task. We use the Air Travel Information System (ATIS) \cite{hemphill-etal-1990-atis} dataset for the task and use the standard dataset splits.

% \subsection{Settings}

\subsection{Datasets}
\paragraph{Named Entity Recognition (NER)} We use the corpora from the CoNLL 2002 and CoNLL 2003 shared tasks \cite{tjong-kim-sang-2002-introduction,tjong-kim-sang-de-meulder-2003-introduction}, which contain four languages in total. We use the standard training/development/test split for experiments.\footnote{\url{https://www.clips.uantwerpen.be/conll2003/ner/}}
\paragraph{Chunking} The chunking datasets are also from the CoNLL 2003 shared task \cite{tjong-kim-sang-de-meulder-2003-introduction} that contains English and German datasets. We use the same standard split as in NER.
\paragraph{Part-Of-Speech (POS) Tagging} Universal Dependencies\footnote{\url{https://lindat.mff.cuni.cz/repository/xmlui/handle/11234/1-2837}} (UD) \cite{11234/1-2837} contains syntactically annotated corpora of over 70 languages. We use universal POS tag annotations with 8 languages for experiments. The list of treebanks is shown in Table \ref{tab:treebank}. We use the standard training/development/test split for experiments.
\paragraph{Slot Filling} Slot filling is a task that interprets user commands by extracting relevant slots, which can be formulated as a sequence labeling task. We use the Air Travel Information System (ATIS) \cite{hemphill-etal-1990-atis} dataset for the task\footnote{We use the same dataset split as \url{https://github.com/sz128/slot_filling_and_intent_detection_of_SLU/tree/master/data/atis-2}.}.
%and split 10\% from the training data as the development set.

\subsection{Settings}
\paragraph{Embeddings} For word embeddings in the NER, chunking and slot filling experiments, we use the same word embedding as in \citet{lample-etal-2016-neural} except that we use \textit{fastText} \cite{bojanowski2017enriching} embedding for Dutch which we find significantly improves the accuracy (more than 5 F1 scores on CoNLL NER). We use \textit{fastText} embeddings for all UD tagging experiments. For character embedding, we use a single layer character CNN with a hidden size of 50, because \citet{yang-etal-2018-design} empirically showed that it has competitive performance with character LSTM. We concatenate the word embedding and character CNN output for the final word representation.

\paragraph{Encoder}
In our experiments, we use three types of encoders. The first is a BiLSTM fed with word and character embeddings, which captures contextual information globally.
The second is a single layer CNN with only word embedding as input, which captures contextual information locally. The third is a single linear layer with word embeddings as input, which does not capture any contextual information. We use these settings for a better understanding of how the decoders perform on each task when the encoders capture different levels of contextual information. 
%Although a single layer CNN loses a significant amount of contextual information, it is much faster than the BiLSTM encoder on GPU.
\paragraph{Decoder}
We use the MaxEnt approach, the traditional CRF approach and AINs with the first-order and factorized second-order CRFs for decoding. We denote these approaches by \textbf{MaxEnt}, \textbf{CRF}, \textbf{AIN-1O} and \textbf{AIN-F2O} respectively. We set the iteration number $M$ to $3$ in AINs because we find that more iterations do not result in further improvement in accuracy. 

\paragraph{Hyper-parameters} For the hyper-parameters, we follow the settings of previous work \cite{akbik-etal-2018-contextual}. We use Stochastic Gradient Descent for optimization with a fixed learning rate of $0.1$ and a batch size of $32$. We fix the hidden size of the CNN and BiLSTM layer to $512$ and $256$ respectively, and the kernel size of CNN to $3$. We anneal the learning rate by 0.5 if there is no improvement in the development sets for $10$ epochs when training. 

\paragraph{Evaluation} We use F1 score to evaluate the NER, slot filling and chunking tasks and use accuracy to evaluate the POS tagging task. We convert the BIO format into BIOES format for NERs, slot filling and chunking datasets and use the official release of CoNLL evaluation script\footnote{\url{https://github.com/chakki-works/seqeval/blob/master/tests/conlleval.pl}} to evaluate the F1 score.

% \paragraph{Datasets} We evaluate our approaches on four tasks: NER, POS tagging, chunking and slot filling. For NER, we use the corpora from the CoNLL 2002 and CoNLL 2003 shared tasks \cite{tjong-kim-sang-2002-introduction,tjong-kim-sang-de-meulder-2003-introduction}. For POS tagging, we use universal POS tag annotations with 8 languages from the Universal Dependencies (UD) \cite{11234/1-2837} dataset. For chunking, we use the corpora from the CoNLL 2003 shared task. We use the Air Travel Information System (ATIS) \cite{hemphill-etal-1990-atis} dataset for slot filling.
% \paragraph{Embedding} For word embeddings in NER, chunking and slot filling experiments, we use the same word embedding as \citet{lample-etal-2016-neural} except that we use \textit{fastText} \cite{bojanowski2017enriching} embedding for Dutch which we find significantly improves the accuracy. We use \textit{fastText} embeddings for all POS tagging experiments. For character embedding, we use a single layer character CNN with a hidden size of 50, as empirically shown by \citet{yang-etal-2018-design} that it has competitive performance with character LSTM. We concatenate the word embedding and character CNN output as the final word representation.

% Please refer to the supplementary materials for more details of the settings.
%We do not experiment with the Viterbi algorithm on the second-order CRFs because it is significantly slower and less frequently used for neural sequence labeling than the other approaches.

\subsection{Results}

\paragraph{Speed}
We report the relative speed improvements over the \textbf{CRF} model based on our PyTorch \cite{NEURIPS2019_9015} implementation run on a GPU server with Nvidia Tesla V100. Following \citet{tsai-etal-2019-small}, we report the training and prediction speed with 10,000 sentences of 32 and 128 words, respectively.
The results (Table \ref{tab:speeds}) show that AINs are significantly faster than \textbf{CRF} in terms of both the full model speed and the decoder speed.
The speed advantage of AINs is especially prominent with long sentences, suggesting their usefulness in tasks like document-level NER.

\paragraph{Accuracy}
We run each approach on each dataset for 5 times and compute its average accuracy.
Because of space limit, we report the accuracy averaged over all the datasets for each task in Table \ref{tab:ner_pos}. Please refer to the supplementary material for the complete results. AINs achieve competitive overall accuracy with \textbf{CRF}, even though AINs take significantly less time than \textbf{CRF}. With the BiLSTM encoder which has the capability to capture global contextual information, AINs achieves almost the same average accuracy as \textbf{CRF}, demonstrating that AINs performing approximate inference with local contextual information are competitive with
%have more advantages than 
\textbf{CRF} with globally exact decoding. With the CNN encoder that encodes local contextual information, AINs are inferior to \textbf{CRF} because both the CNN layer and our approaches utilize only local information. Without any contextual encoders (Word Only), the accuracy of these decoders vary significantly over tasks. For NER and chunking, \textbf{CRF} is the strongest, but our approaches only marginally underperform \textbf{CRF} while significantly outperform \textbf{MaxEnt}. For POS tagging and slot filling, our approaches outperform \textbf{CRF}, which implies that local information might be more beneficial for these tasks. Comparing \textbf{AIN-1O} and \textbf{AIN-F2O}, \textbf{AIN-F2O} is stronger when the encoder is weak, but their performance gap becomes smaller and eventually disappears when the encoder gets stronger.

\subsection{Discussion on Transformers}
Recently, the Transformer \citep{NIPS2017_7181} encoder has significantly improved the performance of tasks such as neural machine translation. The Transformer can be parallelized over the input words while the BiLSTM layer needs sequential computation. However, the transformer structure is rarely applied in sequence labeling tasks. One possible reason is that the performance of models with Transformers encoders are inferior to the performance of models with the BiLSTM encoders. In our experiments, a six-layer transformer with a MaxEnt decoder achieves an F1 score of only 80.00 on CoNLL English NER, which is significantly lower than the 91.00 F1 score of our BiLSTM+MaxEnt model (Table \ref{tab:ner_chunk_sf}). For the speed, the six-layer transformer model with a MaxEnt decoder is 1.58/1.14 times slower than the single-layer BiLSTM model with a MaxEnt decoder with sentences of 32/128 words respectively. Therefore, we do not include the Transformer encoder in our experiments.

% In our implementation, the transformer models with the MaxEnt layer only achieve 80.00 F1 scores in our experiment with 6 layers of transformers, which is significantly lower than our BiLSTM models with the MaxEnt layer (Table \ref{tab:ner_chunk_sf}). For the speed, the six-layer transformer models is 1.58/1.14 times slower than the single-layer BiLSTM models with sentences of 32/128 words respectively.

\section{Conclusion}
In this paper, we propose approximate inference networks (AIN) that use Mean-Field Variational Inference (MFVI) instead of exact probabilistic inference algorithms such as the forward-backward and Viterbi algorithms for training and prediction on the conditional random field for sequence labeling. The MFVI algorithm can be unfolded as a recurrent neural network and connected with the encoder to form an end-to-end neural network. AINs can be parallelized over different positions in the sequence. Empirical results show that AINs are significantly faster than traditional CRF and do very well in tasks that require more local information. Our approaches achieve competitive accuracy on 4 tasks with 15 datasets over three encoder types. Our code is publicly available at \url{https://github.com/Alibaba-NLP/AIN}.
% In future work, we plan to further improve the accuracy of our approaches through knowledge distillation that transfers the knowledge of stronger teacher models such as state-of-the-art sequence labelers with contextual embeddings to our AIN models.
\section*{Acknowledgements}
This work was supported by the National Natural Science Foundation of China (61976139). This work also was supported by Alibaba Group through Alibaba Innovative Research Program. 

\bibliography{anthology,emnlp2020}
\bibliographystyle{acl_natbib}

\appendix
\section{Appendix}

\subsection{Detailed Results}
The detailed results for the four tasks are shown in Table \ref{tab:pos} and \ref{tab:ner_chunk_sf}. We use ISO 639-1 codes\footnote{\url{https://en.wikipedia.org/wiki/List_of_ISO_639-1_codes}} to represent each language.

\begin{table}[t!]

\centering
\begin{tabular}{lr}
\hlineB{4}
Language & Treebank\\
\hline
de & GSD\\
en & EWT\\
es & GSD\\
fr & Sequoia\\
it & PoSTWITA\\
nl & LassySmall\\
sl & SST\\
sv & LinES\\
\hlineB{4}
\end{tabular}
\caption{The list of treebank that we used in UD POS tagging.}
\label{tab:treebank}
\end{table}

\begin{table*}[t]
\setlength\tabcolsep{3pt}
\small
\centering
\begin{tabular}{l||cccccccc|c}
\hlineB{4}
& \multicolumn{9}{c}{\bf \textsc{POS tagging}}\\
\hline
Model & de & en & es & fr & it & nl & sl & sv & avg \\
\hline\hline
\multicolumn{10}{c}{\bf \textsc{Word-Char-BiLSTM}}  \\
\hline
MaxEnt & 94.19$\pm0.04$ & 95.70$\pm0.07$ & \textbf{96.44}$\pm0.05$ & 98.00$\pm0.06$ & 92.76$\pm0.17$ & 95.09$\pm0.13$ & 90.96$\pm0.69$ & 95.56$\pm0.07$ & 94.84 \\
\hline
CRF & \textbf{94.27}$\pm0.11$ & 95.71$\pm0.06$ & 96.37$\pm0.09$ & \textbf{98.06}$\pm0.04$ & 92.87$\pm0.15$ & 95.10$\pm0.17$ & 91.38$\pm1.12$ & 95.55$\pm0.09$ & 94.91 \\
AIN-1 & 94.23$\pm0.06$ & 95.73$\pm0.05$ & 96.39$\pm0.10$ & 98.04$\pm0.10$ & \textbf{93.13}$\pm0.19$ & 95.10$\pm0.20$ & \textbf{91.42}$\pm0.28$ & \textbf{95.69}$\pm0.05$ & \textbf{94.97} \\
AIN-F2 & 94.11$\pm0.22$ & \textbf{95.76}$\pm0.05$ & 96.34$\pm0.05$ & 97.99$\pm0.11$ & 92.87$\pm0.20$ & \textbf{95.24}$\pm0.16$ & 91.38$\pm0.44$ & 95.59$\pm0.07$ & 94.91 \\
\hline\hline
\multicolumn{10}{c}{\bf \textsc{Word CNN}}  \\
\hline
MaxEnt & 92.36$\pm0.19$ & 93.99$\pm0.12$ & 95.91$\pm0.06$ & 97.62$\pm0.05$ & 92.49$\pm0.08$ & 94.51$\pm0.08$ & 91.39$\pm0.18$ & 93.76$\pm0.15$ & 94.00 \\
\hline
CRF & 93.06$\pm0.17$ & \textbf{94.22}$\pm0.10$ & \textbf{96.09}$\pm0.08$ & 97.68$\pm0.07$ & 92.63$\pm0.05$ & 94.63$\pm0.16$ & 91.65$\pm0.23$ & 94.15$\pm0.17$ & 94.26 \\
AIN-1 & \textbf{93.11}$\pm0.14$ & 94.21$\pm0.05$ & 96.02$\pm0.06$ & 97.73$\pm0.05$ & 92.64$\pm0.06$ & 94.58$\pm0.07$ & 91.77$\pm0.20$ & 94.26$\pm0.11$ & 94.29 \\
AIN-F2 & 92.99$\pm0.12$ & 94.17$\pm0.13$ & 96.00$\pm0.04$ & \textbf{97.75}$\pm0.03$ & \textbf{92.69}$\pm0.06$ & \textbf{94.68}$\pm0.04$ & \textbf{91.84}$\pm0.23$ & \textbf{94.47}$\pm0.10$ & \textbf{94.32} \\
\hline\hline
\multicolumn{10}{c}{\bf \textsc{Word Only}}  \\
\hline
MaxEnt & 89.44$\pm0.08$ & 87.57$\pm0.12$ & 93.02$\pm0.05$ & 94.82$\pm0.07$ & 89.23$\pm0.08$ & 91.63$\pm0.17$ & 88.56$\pm0.24$ & 90.01$\pm0.06$ & 90.53 \\
\hline
CRF & 91.55$\pm0.13$ & 91.04$\pm0.22$ & 94.64$\pm0.05$ & 96.65$\pm0.10$ & 91.56$\pm0.05$ & 93.28$\pm0.12$ & 90.02$\pm0.24$ & 93.55$\pm0.09$ & 92.79 \\
AIN-1 & 91.53$\pm0.08$ & 91.47$\pm0.09$ & 94.77$\pm0.05$ & 96.67$\pm0.05$ & 91.62$\pm0.03$ & \textbf{93.46}$\pm0.03$ & 89.65$\pm0.37$ & 93.54$\pm0.10$ & 92.84 \\
AIN-F2 & \textbf{91.75}$\pm0.09$ & \textbf{91.76}$\pm0.12$ & \textbf{94.82}$\pm0.03$ & \textbf{96.95}$\pm0.05$ & \textbf{91.63}$\pm0.06$ & 93.32$\pm0.13$ & \textbf{90.17}$\pm0.23$ & \textbf{93.86}$\pm0.09$ & \textbf{93.03} \\
\hlineB{4}
\end{tabular}
\caption{Averaged accuracy scores on POS tagging.}
\label{tab:pos}
\end{table*}

\begin{table*}[t]
\setlength\tabcolsep{3pt}
\small
\centering
\begin{tabular}{l||cccc|c||cc|c||c}
\hlineB{4}
& \multicolumn{5}{c||}{\bf \textsc{NER}} & \multicolumn{3}{c||}{\bf \textsc{Chunk}} & \bf \textsc{SF}   \\
\hline
Models & de & en & es & nl & avg & de & en & avg & en \\
\hline\hline
\multicolumn{10}{c}{\bf \textsc{Word-Char-BiLSTM}}  \\
\hline
MaxEnt & 75.63$\pm0.23$ & 91.00$\pm0.23$ & 84.53$\pm0.50$ & 83.78$\pm0.38$ & 83.74 & 93.80$\pm0.14$ & 91.36$\pm0.10$ & 92.58 & 95.47$\pm0.06$ \\
\hline
CRF & \textbf{76.46}$\pm0.24$ & 91.14$\pm0.16$ & 85.29$\pm0.36$ & 83.80$\pm0.33$ & 84.17 & \textbf{94.06}$\pm0.07$ & 91.70$\pm0.08$ & \textbf{92.88} & 95.52$\pm0.10$ \\
AIN-1O & 76.34$\pm0.34$ & 91.07$\pm0.10$ & \textbf{85.37}$\pm0.07$ & \textbf{84.12}$\pm0.53$ & \textbf{84.22} & 94.03$\pm0.02$ & \textbf{91.71}$\pm0.05$ & 92.87 & \textbf{95.59}$\pm0.11$ \\
AIN-F2O & 76.17$\pm0.28$ & \textbf{91.22}$\pm0.20$ & 85.30$\pm0.32$ & 83.76$\pm0.57$ & 84.11 & 94.02$\pm0.04$ & 91.69$\pm0.08$ & 92.85 & 95.58$\pm0.14$\\
\hline\hline
\multicolumn{10}{c}{\bf \textsc{Word CNN}}  \\
\hline
MaxEnt & 69.40$\pm0.15$ & 84.86$\pm0.41$ & 70.02$\pm0.62$ & 76.46$\pm0.28$ & 75.19 & 88.29$\pm0.10$ & 85.80$\pm0.65$ & 87.05 & 91.07$\pm0.01$ \\
\hline
CRF & \textbf{71.12}$\pm0.25$ & \textbf{87.58}$\pm0.21$ & \textbf{80.34}$\pm0.58$ & \textbf{78.70}$\pm0.30$ & \textbf{79.44} & \textbf{89.68}$\pm0.21$ & \textbf{88.73}$\pm0.18$ & \textbf{89.21} & 92.24$\pm0.27$ \\
AIN-1O & 70.00$\pm0.28$ & 86.94$\pm0.43$ & 78.95$\pm0.51$ & 77.98$\pm0.38$ & 78.47 & 89.21$\pm0.11$ & 88.51$\pm0.15$ & 88.86 & 92.18$\pm0.14$ \\
AIN-F2O & 70.08$\pm0.92$ & 87.01$\pm0.22$ & 79.80$\pm0.38$ & 77.95$\pm0.47$ & 78.71 & 89.33$\pm0.12$ & 88.16$\pm0.30$ & 88.75 & \textbf{92.26}$\pm0.26$ \\
\hline\hline
\multicolumn{10}{c}{\bf \textsc{Word Only}}  \\
\hline
MaxEnt & 36.24$\pm1.77$ & 63.68$\pm1.08$ & 52.42$\pm1.73$ & 56.73$\pm0.77$ & 52.27 & 81.21$\pm0.33$ & 75.14$\pm0.41$ & 78.17 & 62.93$\pm0.33$ \\
\hline
CRF & 55.10$\pm2.87$ & \textbf{81.76}$\pm0.39$ & \textbf{76.53}$\pm0.80$ & \textbf{75.71}$\pm0.39$ & \textbf{72.28} & \textbf{90.56}$\pm0.24$ & \textbf{88.21}$\pm0.34$ & \textbf{89.39} & 76.82$\pm0.57$ \\
AIN-1O & \textbf{57.25}$\pm2.16$ & 79.68$\pm0.25$ & 70.44$\pm0.72$ & 73.55$\pm0.21$ & 70.23 & 90.04$\pm0.18$ & 87.35$\pm0.29$ & 88.69 & 88.76$\pm0.65$ \\
AIN-F2O & 56.36$\pm5.97$ & 81.16$\pm0.37$ & 73.03$\pm1.86$ & 74.09$\pm0.24$ & 71.16 & 90.04$\pm0.15$ & 87.56$\pm0.24$ & 88.8 & \textbf{88.86}$\pm0.41$ \\
\hlineB{4}
\end{tabular}
\caption{Averaged F1 scores on NER, chunking and slot filling for each language. \textbf{SF} represents the slot filling task. $\star$: for reference.}
\label{tab:ner_chunk_sf}
\end{table*}

\end{document}